\documentclass{article}
\usepackage{spconf,amsmath,graphicx,url}
\usepackage{times}
\usepackage{latexsym}
\usepackage{multirow}

\title{General Phrase Debiaser: Debiasing Masked Language Models at a Multi-Token Level}
\name{Bingkang Shi$^{1,2}$, Xiaodan Zhang$^{1}$, Dehan Kong$^{3}$, Yulei Wu$^{4}$, Zongzhen Liu$^{1}$, Honglei Lyu$^{1}$, Longtao Huang$^{3}$}
  
\address{$^{1}$ {Chinese Academy of Sciences, Institute of Information Engineering, Beijing 100093, China} \\
      $^{2}$ {School of Cyber Security, University of Chinese Academy of Sciences, Beijing, China} \\
      $^{3}$ {Alibaba Group, Alibaba Artificial Intelligence Governance Laboratory, Beijing, China} \\
      $^{4}$ {University of Bristol, Department of Electrical and Electronic Engineering, Bristol, BS8 1UB, UK}
      }

\begin{document}
%
\maketitle
\begin{abstract}
The social biases and unwelcome stereotypes revealed by pretrained language models are becoming obstacles to their application. Compared to numerous debiasing methods targeting word level, there has been relatively less attention on biases present at phrase level, limiting the performance of debiasing in discipline domains. In this paper, we propose an automatic multi-token debiasing pipeline called \textbf{General Phrase Debiaser}, which is capable of mitigating phrase-level biases in masked language models. Specifically, our method consists of a \textit{phrase filter stage} that generates stereotypical phrases from Wikipedia pages as well as a \textit{model debias stage} that can debias models at the multi-token level to tackle bias challenges on phrases. The latter searches for prompts that trigger model's bias, and then uses them for debiasing. State-of-the-art results on standard datasets and metrics show that our approach can significantly reduce gender biases on both career and multiple disciplines, across models with varying parameter sizes. 
\end{abstract}
\section{Introduction}
\label{sec:intro}

Recently,  masked language models (MLMs) \cite{devlin2018bert, lanalbert, liu2019roberta, NEURIPS2019_dc6a7e65, sanh2019distilbert} are employed in both traditional tasks like text classification and diverse multimodal tasks when combined with models like image generators \cite{radford2021learning, rombach2022high}. We aim to develop MLMs with minimal human biases, even when the pretraining data unavoidably contains these biases. However, correcting implicit biases in pretrained MLMs can be very challenging, especially considering the high cost of retraining models from scratch.

Existing studies \cite{liang2020towards, wolfe2021low, kaneko2022debiasing, ct1965, webster2020measuring, chengfairfil} have introduced intuitive approaches that use additional corpus to retrieve contextualized embeddings or locate the biases and fine-tune accordingly. But they are rely on external human-written corpus. Auto-Debias\cite{guo2022auto} hires the prompt\cite{jiang2020can} template "[attribute word] [T]...[T] [MASK]" to guide MLMs to automatically search for prompts that makes the model show its bias, and then fine-tune MLMs with them. Nevertheless, real-world language environments are not so ideal, meaning both attribute words and stereotypes should be treated as multi-token. While these method only correct biases at the word level, lead to struggling at the phrase level. 

Motivated by this, we propose an automatic multi-token debias pipeline called \textbf{General Phrase Debiaser} to address the limitations of automatic debiasing mentioned above. The major contributions of our work are:
\begin{itemize}
	\item Unlike existing methods, we debias MLMs at the phrase granularity. In order to reduce the cost of manually constructing the phrase list, we get the stereotypical phrases filtered from hyperlinks of \textbf{Wikipedia} pages in \textit{Phrase Filter Stage}. 
	\item With the \textit{multi-token debias head} we proposed, “discriminative” prompts can be searched in \textit{Model Debias Stage}. These cloze-style prompts have the highest disagreement in generating stereotypical phrases (e.g., mathematical theory/dance art) with respect to demographic words (e.g., man/woman). Then we fine-tune the model using searched prompts.
	\item Different from the Auto-Debias' fine-tuning stage, our approach derives loss from stereotypical phrases, rather than from the entire vocabulary belonging to the model itself. This allows our method to adjust the model parameters more specifically without affecting any other gender-independent word or knowledge. 
	\item We conduct experiments on three well-known open-source MLMs: BERT\cite{devlin2018bert}, ALBERT\cite{lanalbert}, and DistilBERT\cite{sanh2019distilbert},  and achieves state-of-the-art performance (\textbf{0.12}, \textbf{0.16}, and \textbf{52}) on SEAT test.
\end{itemize}

Our code and debiased model files are available at https://\\github.com/BingkangShi/general-phrase-debiaser.

\begin{figure*}[htb]
\centering
\resizebox{\textwidth}{!}{\includegraphics[width=1\textwidth]{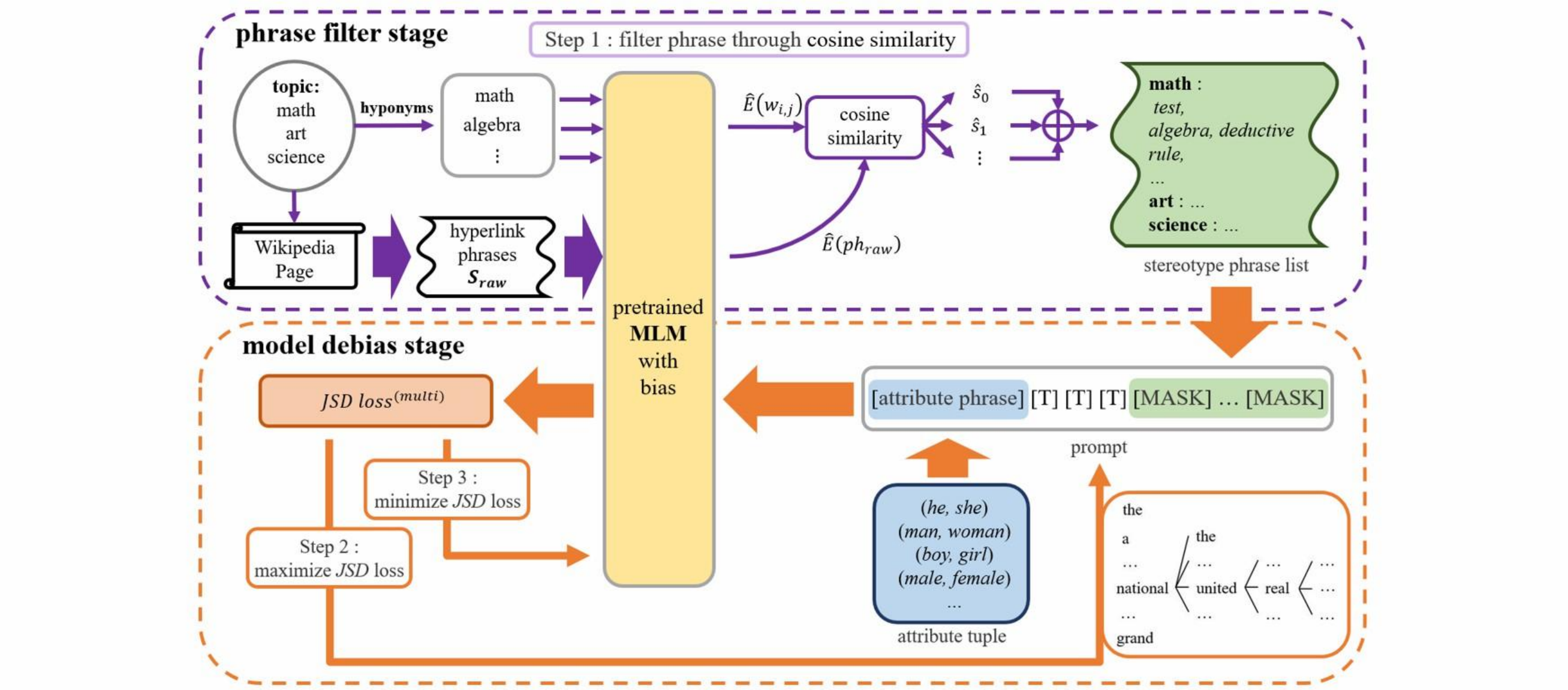}}
\caption{\label{frame}
The proposed General phrase Debiaser pipeline has two stages. In \textit{phrase filter stage}, we filter out stereotypical phrases $S_{weighted}$ and $S_{unweighted}$ from WikiPedia pages with MLM and stereotypical seeds. In \textit{model debias stage}, we search biased prompts at multi-token granularity, and fine-tuning MLM with them.}
\end{figure*}

\begin{figure}[htb]
\centering
\resizebox{\linewidth}{!}{
\includegraphics[width=1\textwidth]{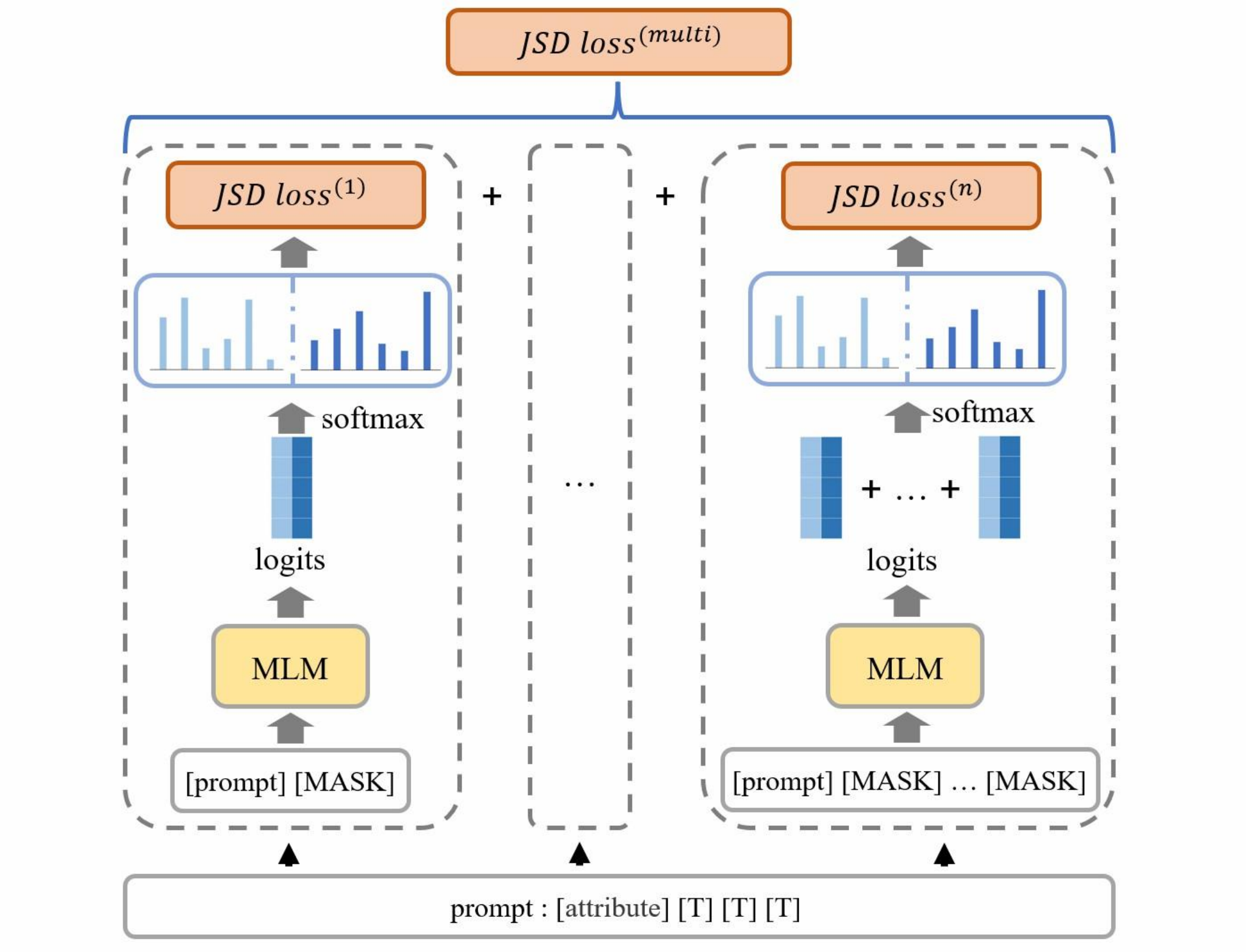}
}
\caption{\label{fig:multi-token}Computation of multi-token JSD loss.}
\end{figure}

\section{General Phrase Debiaser}
\label{sec:GPD}

\subsection{Phrase Filter Stage}
\label{ssec:filter}

To minimize the cost of manually constructing stereotypes in many specific domains, we use the MLM which needs to be debiased to filter hyperlinks of WikiPedia pages. Hyperlinked phrases $S_{raw}$ that semantically similar to stereotypical seeds can be filtered out. And stereotype seeds comprising $N_{topic}$ topics need to be manually specified, with each topic having $ n_i$ hyponyms. We choose career, math, art, and science as topics to construct stereotypes, so $N_{topic}$ is 4 in this paper. It should be noted that the filtered phrases under the math, art, and science topics are generated by our \textit{Phrase Filter Stage}, while the phrases under the career topic were provided by previous work \cite{kaneko2021debiasing}.

Let $\mathcal{M}$ be a MLM and $\mathcal{M}_{CLS}$ be the process of computing the classification embedding (CLS) that represents a sentence. The embedding of a phrase can be computed as follows:
\begin{equation}
\hat{E}\left(x\right)=\frac{1}{\left|T\right|}\sum_{t\in T}{\mathcal{M}_{CLS}\left(t\left(x\right)\right)}\\
\end{equation}
where $t$ represents a sentence template, and $T$ is a set that includes all templates. We refer to the 14 blank-filling templates used in the SEAT test \cite{may2019measuring}, such as "this is a \_\_." or "\_\_ is here.".

Then the cosine similarity between phrase $ph_{raw}\in S_{raw}$ and $w_{i,j}$ can be computed through:
\begin{equation}
d\left(i,j,ph_{raw}\right)=cos(\hat{E}\left(ph_{raw}\right),\ \hat{E}\left(w_{i,j}\right))\\
\end{equation}
where $w_{i,j}$ is the $j$-th hyponym of the $i$-th topic, $j\in n_i$ and $i\in N_{topic}$. 

We define the quantity of $ph_{raw}$ in phrase set ${\hat{s}}_{i,j}$ as $topK_p$, according to $d\left(i,j,ph_{raw}\right)$ sorted by ascending order. So we can collect $S_{weighted}$ with:
\begin{equation}
S_{weighted}=(S_0,S_1,\cdots,{\ S}_{N_{topic}-1}\ )\\
\end{equation}
where the phrase set of $i$-th topic is $S_i=({\hat{s}}_0,{\hat{s}}_1,\cdots,\ {\hat{s}}_{n_i-1}\ )$. After removing duplicate phrases, $S_{weighted}$ is transformed into $S_{unweighted}$.

\subsection{Finding Biased Prompts}
\label{ssec:bias_search}

Previous attempts of Auto-Debias \cite{guo2022auto} used cloze-style prompts to detect biases in attribute words within stereotypes. Let $\mathcal{V}$ be vocabulary of a MLM, and a prompt $x_{prompt}\in\mathcal{V}$ is a sequence of words with one masked token [MASK] and one attribute token. A MLM can be probed by a cloze-style prompt, such as $x_{prompt}$="[attribute] majors in [MASK].". The "[attribute]" is assigned to be filled in a set $\mathcal{C}=\{(c_{1,1},c_{1,2},\cdots,c_{1,m}),(c_{2,1},c_{2,2},\cdots,c_{2,m}),\cdots\}$ composed of $m$-tuples, derived from the gender word list in \cite{kaneko2021debiasing}. And the position of "[MASK]" serves the purpose of being predicted by $\mathcal{M}$ for a stereotypical word. So we can obtain stereotypical word probability as:
\begin{equation}
\begin{aligned}
&P\left([MASK]=v|\mathcal{M},x_{prompt}\left(c\right)\right)\\
&=softmax\left(\mathcal{M}_{[MASK]}\left(v|x_{prompt}\left(c\right)\right)\right)\\
\end{aligned}
\end{equation}
where $v\in\mathcal{V}$. $x_{prompt}\left(c\right) = c \oplus x \oplus \left\lbrack {MASK} \right\rbrack$ is a string composed of $x_{prompt}$ and $c \in \mathcal{C}$. For example, $x_{prompt}\left(c\right)$ = "$c$ majors in [MASK].".

While the above method is effective only when stereotypes are single-token. Thus we introduce a probability calculation method for stereotypes at multi-token granularity (as shown in Fig.2):
\begin{equation}\label{eq:5}
\begin{aligned}
&P_{c_{k,m}}^{(n)} = P( 
\left\lbrack \mathbf{t}\mathbf{a}\mathbf{r}\mathbf{g}\mathbf{e}\mathbf{t} \right\rbrack = {ph}_{i} |\mathcal{M},x_{prompt}^{'}(c_{k,m}^{({n})}) )\\
&= softmax ({\sum\limits_{l\in n}{logit_{l}}})\\
\end{aligned}
\end{equation}
while the logit corresponding to the [MASK] token position should be:
\begin{equation}
logit_{l} = \mathcal{M}_{{\lbrack MASK\rbrack}_{l}} ( {ph}_{i} | x_{prompt}^{'}(c_{k,m}^{({n})}) )\\
\end{equation}
where $\left\lbrack \mathbf{t}\mathbf{a}\mathbf{r}\mathbf{g}\mathbf{e}\mathbf{t} \right\rbrack$ is multiple [MASK] token sequence of length $n$, and $n$ is the maximum length of stereotypical phrase ${ph}_{i}\in S_{unweighted}$. $c_{k,m}^{({n})}$ containing $n$ [MASK] tokens is the $m$-th phrase of the $k$-th tuple in set $\mathcal{C}$. Here $x_{prompt}\left(c\right)$ should evolve into:
\begin{equation}
\begin{aligned}
&x_{prompt}^{'}\left( c_{k,m}^{({n})} \right) = c_{k,m}^{({n})} \oplus x \oplus \left\lbrack {\mathbf{t}\mathbf{a}\mathbf{r}\mathbf{g}\mathbf{e}\mathbf{t}} \right\rbrack\\
&= c_{k,m}^{({n})} \oplus x \oplus \left\lbrack {MASK} \right\rbrack\left\lbrack {MASK} \right\rbrack\cdots\left\lbrack {MASK} \right\rbrack\\
\end{aligned}
\end{equation}

By repeatedly applying Eq. (\ref{eq:5}), we can obtain the distributions $P^{(1)}$, $P^{(2)}$, ... , $P^{(n)}$ for stereotypical phrases with different token length. In step 2 and step 3 in Fig.1, we use Jensen-Shannon Divergence (JSD), which is a symmetric and smooth Kullback–Leibler divergence (KLD), to measure the difference between multiple distributions $P_{c_{k,1}}^{(n)} + P_{c_{k,2}}^{(n)} + \cdots + P_{c_{k,m}}^{(n)}$ as follow: 
\begin{equation}
\begin{aligned}
&{JSD\ loss}_{k}^{({n})} = {\sum\limits_{n}{JSD ( {P_{c_{k,1}}^{(n)},P_{c_{k,2}}^{(n)},\cdots,P_{c_{k,m}}^{(n)}} )}}\\
&= \frac{1}{m}{\sum\limits_{i}{KLD\left( P_{c_{k,i}}^{(n)} \middle| \middle| \frac{P_{c_{k,1}}^{(n)} + P_{c_{k,2}}^{(n)} + \cdots + P_{c_{k,m}}^{(n)}}{m} \right)}}\\
\end{aligned}
\end{equation}
In this paper, JSD measures the difference between the two-gender distributions, so $m=2$. The KLD between two distributions $p_{i}$ and $p_{j}$ can be computed as: $KLD\left( p_{i} \middle| \middle| p_{j} \right) = {\sum_{v \in \mathcal{V}}{p_{i}(v)}}{\mathit{\log}( \frac{p_{i}(v)}{p_{j}(v)} )}$.

The loss of an input $x_{prompt}^{'}$ can be seen as the sum of the overall probability distribution differences of $\mathcal{C}$:
\begin{equation}
\begin{aligned}
&loss\left( x_{prompt}^{'} \right) = {\sum\limits_{k}{JSD\ loss}_{k}^{({multi})}} \\
&= {\sum\limits_{k}{\sum\limits_{n}{JSD\ loss}_{k}^{(n)}}}\\
\end{aligned}
\end{equation}

Step 2 in Fig.1 shows we employ Beam Search\cite{freitag2017beam} to find biased prompts that maximize the $loss ( x_{prompt}^{'} )$. Searched biased prompts will be collected for fine-tuning MLM in the step 3.

\subsection{Fine-tuning MLM with Prompts}
\label{ssec:bias_search}

\begin{table*}[htb]
\centering
\resizebox{0.45\textheight}{!}{
\begin{tabular}{lccccccc}
\hline
\textbf{Model} & \textbf{SEAT-6} & \textbf{SEAT-6b} & \textbf{SEAT-7} & \textbf{SEAT-7b} & \textbf{SEAT-8} & \textbf{SEAT-8b} & \textbf{avg.}\\
\hline
BERT & 0.48 & 0.11 & 0.25 & 0.25 & 0.40 & 0.61 & 0.35 \\
+Context-Debias\cite{kaneko2021debiasing} & 1.13 & - & 0.34 & - & 0.12 & - & 0.53 \\
+FairFil\cite{chengfairfil} & 0.18 & 0.08 & \textbf{0.12} & \textbf{0.08} & 0.20 & 0.24 & 0.15 \\
+Auto-Debias\cite{guo2022auto} & 0.08 & \textbf{0.02} & 0.36 & 0.40 & 0.12 & 0.20 & 0.20 \\
+General Phrase Debiaser  & \textbf{0.00} & 0.13 & 0.19 & 0.10 & \textbf{0.02} & 0.27 & \textbf{0.12} \\
\hline
ALBERT & 0.51 & \textbf{0.02} & 0.58 & 1.02 & 0.99 & 1.20 & 0.72 \\
+Context-Debias\cite{kaneko2021debiasing} & 0.18 & - & 0.05 & - & 0.77 & - & 0.33 \\
+General Phrase Debiaser & \textbf{0.04} & 0.30 & \textbf{0.01} & \textbf{0.02} & \textbf{0.33} & \textbf{0.29} & \textbf{0.16} \\
\hline
DistilBERT & 1.26 & \textbf{0.25} & 0.31 & 1.22 & 0.74 & 0.98 & 0.79 \\
+Context-Debias\cite{kaneko2021debiasing} & 1.34 & - & 1.01 & - & 0.97 & - & 1.11 \\
+General Phrase Debiaser & \textbf{0.60} & 0.32 & \textbf{0.21} & \textbf{0.99} & \textbf{0.23} & \textbf{0.79} & \textbf{0.52} \\
\hline
\end{tabular}
}
\caption{\label{main_evaluation}
Gender debiasing results of SEAT on BERT, ALBERT and DistilBERT. SEAT-6 and SEAT-6b are tests about career, SEAT-7, SEAT-7b, SEAT-8 and SEAT-8b are about discipline. Scores closer to 0 are better. "-" means the value is not reported in the original paper.
}
\end{table*}
\begin{table*}[htb]
\centering
\resizebox{0.60\textheight}{!}{
\begin{tabular}{lccccccccc}
\hline
\textbf{Model} & \textbf{CoLA} & \textbf{SST-2} & \textbf{MRPC} & \textbf{STS-B} & \textbf{QQP} & \textbf{MNLI} & \textbf{QNLI} & \textbf{RTE} & \textbf{WNLI}\\
\hline
BERT & 0.59 & 0.93 & 0.89/0.85 & 0.89/0.88 & 0.91/0.88 & 0.85/0.85 & 0.92 & 0.65 & 0.56\\
+General Phrase Debiaser & 0.56 & 0.93 & 0.89/0.84 & 0.89/0.89 & 0.90/0.88 & 0.85/0.85 & 0.92 & 0.65 & 0.56\\
\hline
ALBERT & 0.55 & 0.92 & 0.92/0.89 & 0.91/0.91 & 0.91/0.88 & 0.85/0.85 & 0.92 & 0.73 & 0.39 \\
+General Phrase Debiaser & 0.54 & 0.93 & 0.90/0.86 & 0.91/0.91 & 0.91/0.88 & 0.85/0.85 & 0.92 & 0.73 & 0.42\\
\hline
DistilBERT & 0.47 & 0.91 & 0.89/0.84 & 0.86/0.86 & 0.90/0.87 & 0.82/0.82 & 0.88 & 0.58 & 0.56\\
+General Phrase Debiaser & 0.46 & 0.91 & 0.89/0.84 & 0.86/0.86 & 0.90/0.87 & 0.82/0.82 & 0.89 & 0.62 & 0.56\\
\hline
\end{tabular}
}
\caption{\label{GLUE_test}
GLUE test results of the original and the gender-debiased MLMs with our method.
}
\end{table*}

Given that existing work \cite{kaneko2021debiasing} demonstrated the presence of biases in all parameters of the models, we choose to fine-tune the entire $\mathcal{M}$ to mitigate biases in the model with searched biased prompts in step 2. This corresponds to step 3 as illustrated in Fig.1. 

In contrast to the prompt search phase, during the debiasing fine-tuning, we aim to minimize $loss( x_{prompt}^{'} )$ to reduce the distribution discrepancy of $\mathcal{M}$ on $S_{weighted}$ induced by $x_{prompt}^{'}$. This distribution discrepancy is specific to $S_{weighted}$, indicating that our method propagates gradients through each phrase in the entire stereotype, rather than debiasing the entire vocabulary $\mathcal{V}$ of $\mathcal{M}$ as done in Auto-Debias. As a result, our debiasing approach pays more attention to stereotypical phrases in $S_{weighted}$ and has less impact on unrelated words.

\section{Results and Evaluation}
\label{Evaluation}

\subsection{Evaluation Data And Details}
\label{details}

\textbf{Debias Data \& Language Capability Data}: We evaluate the proposed General Phrase Debiaser on two dataset: (1) the \textbf{Sentence Embedding Association Test (SEAT)} \cite{may2019measuring} which provides a commonly used metric for assessing biases in PLM embeddings, and (2)the \textbf{General Language Understanding Evaluation (GLUE)} benchmark \cite{wangglue} which measures common language modeling capability. We evaluate our method on 3 MLMs with differnt sized parameters: BERT \cite{devlin2018bert}, ALBERT \cite{lanalbert}, and distilBERT \cite{sanh2019distilbert}, and compare the proposed method with 3 other algorithms: Context-Debias\cite{kaneko2021debiasing}, FairFil\cite{chengfairfil} as well as Auto-Debias\cite{guo2022auto}.

\textbf{Implementation Details}: In step 2 of Fig.1, we use $S_{unweighted}$ which has 624 phrases. The maximum biased prompt length $PL$ is 5 and beam search width $K$ is 100. We use the 5,000 highest frequency words in Wikipedia as the search space $\mathcal{V}$, to avoid noise in the vocabulary $\mathcal{V}$ and speed up the prompt search process. In step 3, we use $S_{weighted}$ with more than 500 phrases instead of $S_{unweighted}$ because $S_{weighted}$ takes into account the varying weights of different stereotypical phrases, resulting in better debiasing effects (as shown in Table 1). And we choose $\mathcal{C}^{*}$ (an extension of $\mathcal{C}$, derived from the gender word list in \cite{kaneko2021debiasing}) as attribute phrases to construct more fine-tuning data. All models are trained with AdamW \cite{loshchilov2017decoupled} optimizer and early stopping strategy. Our experiments run on a single NVIDIA 3090Ti.

\subsection{Evaluation Result And Analysis}
\label{result}

We run General Phrase Debiaser in both career field and discipline field at the same time. The effect size score of the SEAT[16] benchmark we report in Table \ref{main_evaluation} measures the association between two sets of target concepts and two sets of attributes. It is obtained by calculating the normalized distance between a set of attribute sentence vectors and a set of concept sentence vectors output by the model, and the closer this distance is to 0, the less biased the model is. The result demonstrates that our method is capable of reducing model biases, lowering original average scores of BERT, ALBERT, and DistilBERT in the six SEAT tests from 0.35, 0.72, and 0.79, respectively, to 0.12, 0.16, and 0.52. Furthermore, compared to other approaches in the entire benchmark, including those relying on manual datasets or generating data automatically, General Phrase Debiaser shows the state-of-the-art debiasing performance. We find the global superiority of our model is derived from three aspects through analysis:

\textbf{Simultaneous Debiasing across Multiple Domains.} Our method can effectively eliminate the gender bias in career, math, art, and science simultaneously, without requiring multiple debiasing process on the same model.

\textbf{Knowledge Debiasing in Phrase Granularity.} Our method operates at the phrase granularity rather than the word granularity, making it easier to probe and mitigate biases in disciplines. For example, in Table 2, General Phrase Debiaser achieves the best average score in the four tests (SEAT-7 to SEAT-8) concerning math, art and science.

\textbf{Keep Language Capability after Debias.}we test gender-debiased versions of BERT, ALBERT, and DistilBERT on the General Language Understanding Evaluation (GLUE) benchmark \cite{wangglue}. The test results are presented in Table \ref{GLUE_test}. Gender-debiased versions of BERT, ALBERT, and DistilBERT show a little decrease in scores compared to the original models on the GLUE test, demonstrating our General MLM Debiaser alleviates the bias concerns while also maintaining language modeling capability.

\section{Conclusion}
\label{Conclusion}

Our proposed method can debias MLMs at phrase granularity while also maintaining language modeling capability, and gets state of the art in SEAT test. Although decoder-only LLMs are gaining popularity, we still consider bias mitigation in encoder-only models crucial. Moreover, the concepts presented here can also be applied to cross-modal models involving encoder-only models.

\section*{Acknowledgements}
This work was supported by Grant 2020YFB1005400 from the National Key R\&D Program of China.


\vfill\pagebreak
\bibliographystyle{IEEEbib}
\bibliography{strings,refs}

\end{document}